\pdfoutput=1
\documentclass[11pt]{article}

\usepackage[preprint]{acl}

\usepackage{times}
\usepackage{latexsym}

\usepackage[T1]{fontenc}
\usepackage[utf8]{inputenc}

\usepackage{microtype}

\usepackage{inconsolata}

\usepackage{times}
\usepackage{latexsym}
\newcommand{\minisection}[1]{\noindent{\bf #1}\hspace{0.6em}}

\usepackage{graphicx}
\usepackage{enumitem}
\usepackage{url}
\usepackage{booktabs}
\usepackage{makecell}

\usepackage{amsmath}
\usepackage{multirow}
\usepackage{lipsum}  
\usepackage{amssymb}
\usepackage{pifont}
\usepackage{adjustbox}
\usepackage{listings}
\newcommand{\cmark}{\ding{51}}
\newcommand{\xmark}{\ding{55}}

\newcommand*\samethanks[1][\value{footnote}]{\footnotemark[#1]}
\usepackage{tikz}
\usepackage{pgfplots}
\usepackage{xspace}
\usepackage{ulem}

\newcommand{\ours}{\textsc{Proc2Pddl}\xspace}

\newcommand{\df}{$\mathbb{DF}$\xspace}
\newcommand{\pf}{$\mathbb{PF}$\xspace}
\newcommand{\dfs}{$\mathbb{DF}$s\xspace}
\newcommand{\pfs}{$\mathbb{PF}$s\xspace}
\newcommand{\txt}{$\mathbb{T}$\xspace}

\usepackage{soul}

\newcommand{\squishlist}{
  \begin{list}{$\bullet$}
    { \setlength{\itemsep}{0pt}      \setlength{\parsep}{3pt}
      \setlength{\topsep}{3pt}       \setlength{\partopsep}{0pt}
      \setlength{\leftmargin}{1.5em} \setlength{\labelwidth}{1em}
      \setlength{\labelsep}{0.5em} } }
\newcommand{\reallysquishlist}{
  \begin{list}{$\bullet$}
    { \setlength{\itemsep}{0pt}    \setlength{\parsep}{0pt}
      \setlength{\topsep}{0pt}     \setlength{\partopsep}{0pt}
      \setlength{\leftmargin}{0.2em} \setlength{\labelwidth}{0.2em}
      \setlength{\labelsep}{0.2em} } }

 \newcommand{\squishend}{
     \end{list} 
 }

\renewcommand{\cite}{\citep}

\definecolor{lightgray}{gray}{0.9}
\definecolor{Box1Color}{RGB}{227, 236, 246}
\definecolor{Box2Color}{RGB}{248, 220, 225}
\definecolor{Box3Color}{RGB}{255, 238, 224}
\definecolor{cbBlue}{RGB}{0, 114, 178}
\definecolor{cbOrange}{RGB}{240, 228, 66}
\definecolor{cbGreen}{RGB}{0, 158, 115}
\definecolor{cbRed}{RGB}{213, 94, 0}
\definecolor{cbPurple}{RGB}{204, 121, 167}
\definecolor{cbSkyBlue}{RGB}{86, 180, 233}
\definecolor{cbGray}{RGB}{128, 128, 128}
\definecolor{CBF1}{RGB}{255,99,132}  % Red
\definecolor{CBF2}{RGB}{54,162,235}  % Blue
\definecolor{CBF3}{RGB}{255,206,86}  % Yellow
\definecolor{CBF4}{RGB}{75,192,192}  % Green-blue
\definecolor{CBF5}{RGB}{153,102,255} % Purple
\definecolor{CBF1b}{RGB}{205,89,112}  % Darker Red
\definecolor{CBF2b}{RGB}{44,142,215}  % Darker Blue
\definecolor{CBF5b}{RGB}{133,92,225}  % Darker Purple

\title{\ours : Open-Domain Planning Representations from Texts}

\author{Tianyi Zhang$^{1}$\thanks{Equal contribution.} \quad
  Li Zhang$^1$\samethanks \quad
  Zhaoyi Hou$^{3}$ \\
  \textbf{Ziyu Wang}$^{1}$ \quad
  \textbf{Yuling Gu}$^2$ \quad
  \textbf{Peter Clark}$^{2}$ \quad \\
  \textbf{Chris Callison-Burch}$^{1}$ \quad
  \textbf{Niket Tandon}$^2$ \\
  $^1$University of Pennsylvania\quad\quad $^2$Allen Institute for Artificial Intelligence \\
  $^3$University of Pittsburg \\
  {\tt \{zty|zharry|ccb\}@upenn.edu}
}

\begin{document}
\maketitle
\begin{abstract}
Planning in a text-based environment continues to be a significant challenge for AI systems. Recent approaches have utilized language models to predict planning domain definitions (e.g., PDDL) but have only been evaluated in closed-domain simulated environments. To address this, we present \ours, the first dataset containing open-domain procedural texts paired with expert-annotated PDDL representations. Using this dataset, we evaluate the task of predicting domain actions (parameters, preconditions, and effects). We experiment with various large language models (LLMs) and prompting mechanisms, including a novel instruction inspired by the zone of proximal development (ZPD), which reconstructs the task as incremental basic skills. Our results demonstrate that \ours is highly challenging for end-to-end LLMs, with GPT-3.5's success rate close to 0\% and GPT-4o's 38\%. With ZPD instructions, GPT-4o's success rate increases to 45\%, outperforming regular chain-of-thought prompting's 34\%. Our analysis systematically examines both syntactic and semantic errors, providing insights into the strengths and weaknesses of language models in generating domain-specific programs.\footnote{Our resources can be found at \url{https://github.com/zharry29/proc2pddl}.}
\end{abstract}
% We hope this analysis and dataset helps future progress towards integrating the best of LMs and formal planning.
%\footnote{Our resources can be found at \url{https://github.com/zharry29/nl-to-pddl-action}.}

\section{Introduction}
\label{sec:intro}

Planning is the task of finding a sequence of actions to achieve a goal in a given environment \cite{fikes1971strips,lavalle2006planning}. In real life, the environment is often described with natural language texts. To enable text-based, automated planning, recent work has used language models (LMs) to \textit{generate plans} \cite{valmeekam2023planbench,stein2023autoplanbench}. However, this approach is found to fall short with regard to both performance and interpretability \cite{valmeekam2023large,valmeekam2023planning}. Alternatively, another recent line of worked has instead used LMs to \textit{translate} the natural language description of environments to planning domain definition language (PDDL) \cite{aeronautiques1998pddl}. This symbolic representation can then be solved by a planner in a plan \cite{collins2022structured,lyu2023faithful,liu2023llmp,xie2023translating,wong2023learning}. Despite of the success of such a neurosymbolic method, all the above work has only been evaluated in \textbf{closed-domains} simulated environments such as a household (e.g., ALFRED \cite{shridhar2020alfred}) or discrete object placement (e.g., BlocksWorld \cite{valmeekam2024planbench}) (as shown in Table~\ref{tab:related_work}). 

\begin{figure}[t]
    \centering
    \includegraphics[width=0.46\textwidth]{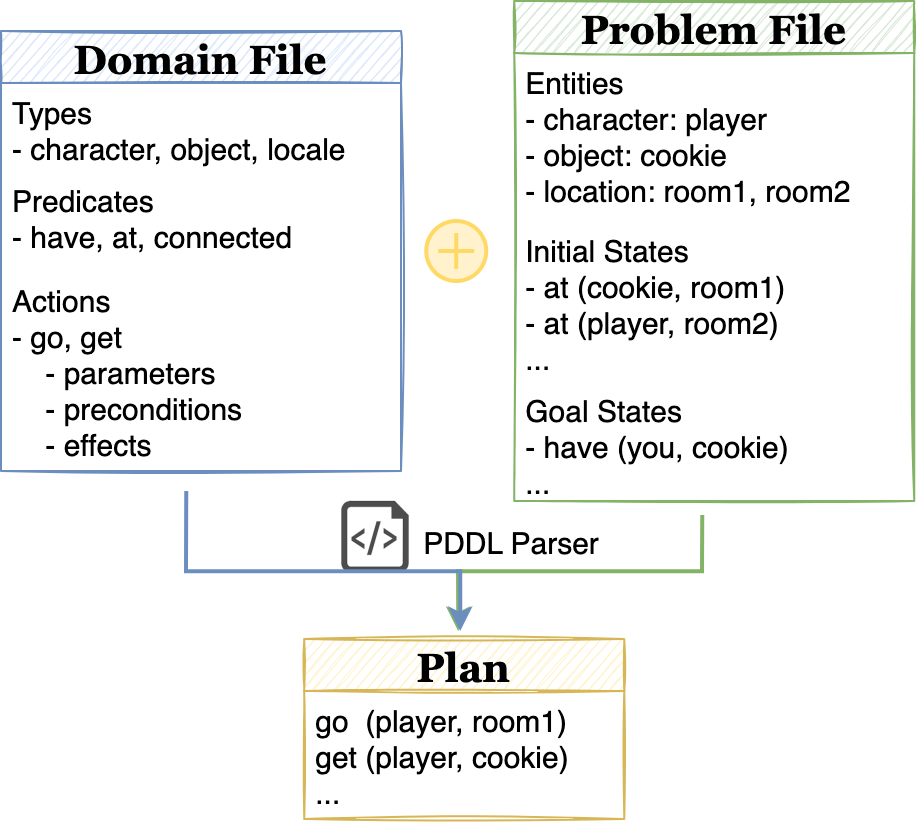}
    \caption{A PDDL solver produces a plan based on a minimal domain file and problem file. Previous work assumes the domain file as given, while we predict the action definitions in the domain file.}
    \vspace{-2ex}
    \label{fig:pddl_example}
\end{figure}

\begin{figure*}[t!]
    \centering
    \includegraphics[width=0.96\textwidth]{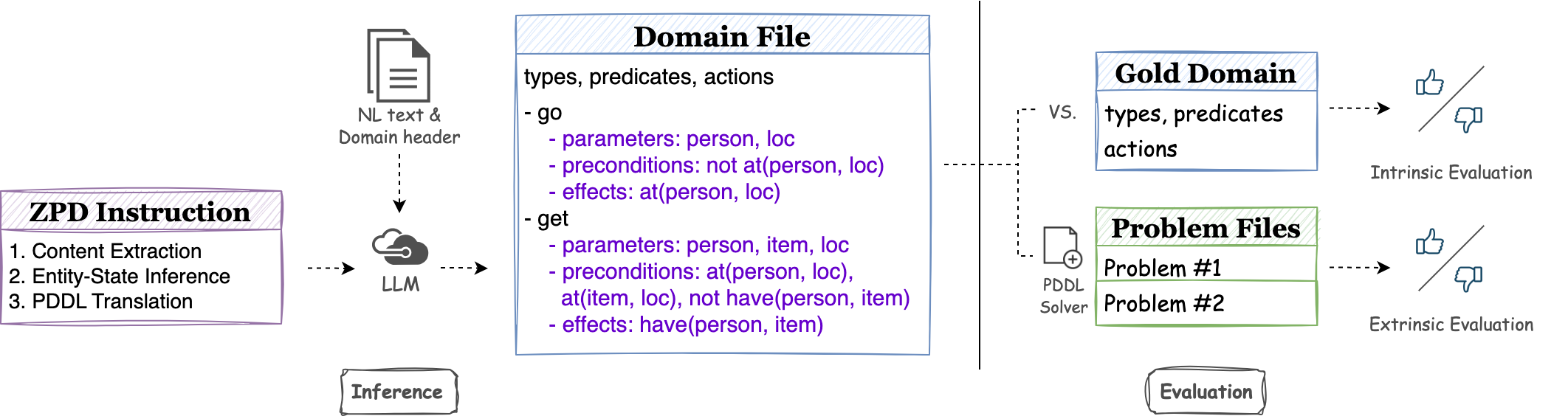}
    \caption{Our formulation of the \df action prediction task is as follows: given a natural language procedure text and a domain file header, a language model (LM) follows Zone of Proximal Development (ZPD) instructions in three sequential skills to predict domain actions, including parameters, preconditions, and effects. During evaluation, the predicted \df is compared to a gold reference and used to solve corresponding \pfs.}
    \label{fig:formulation}
\end{figure*}

% On the other hand, generative large language models (LMs) can provide fluent and sensible open-domain \textit{textual} plans such as a recipes or step-by-step instructions \cite{sakaguchi-etal-2021-proscript-partially,lyu-etal-2021-goal}, catered to human users instead of agents. 

To enable \textbf{open-domain}, text-based planning, we propose \ours, a dataset to evaluate models' ability to generate PDDL given procedural texts. \ours consists of 27 pairs of open-domain procedures and PDDL representations. Each PDDL representation include a domain file \df that models the types, predicates, and actions, and a problem file \pf that models the entities, initial states, and goal states, as illustrated in Figure~\ref{fig:pddl_example}. Because \ours is not bound to any simulation, the PDDL representations are manually annotated by experts trained on this task to ensure validity, resulting in 27 domain files and 95 problem files. 

Using this dataset, we study the task of action modeling \cite{lindsay2017framer} formulated as follows. The input is some relevant natural language texts and the \textit{header} of a \df (i.e., types, predicates, and names of actions). Based on a ZPD instruction, the output is the \textit{domain actions} in the \df (i.e., parameters, preconditions, and effects). During evaluation, the predicted \df is 1) compared to a ground-truth \df as intrinsic evaluation, and 2) provided to a PDDL solver with ground-truth \pfs for the existence and correctness of plans as extrinsic evaluation. Our system is delineated in Figure~\ref{fig:formulation}. 
% We then evaluate the existence and correctness of such a plan. 
In this formulation, our assumption of the \df header is necessary to ensure the consistency of semantics between the \df and the \pf for evaluation. It is also empirically motivated; for example, a kitchen robot may have access to the types like `ingredients' and predicates like `diced' via some information extraction system given descriptive texts, but it may still need to predict, for ``swinging a knife'', the precondition that it is only safe to do so to the `ingredients' and the effect that they will become `diced'. 

Through our experiment, we show that the task of action modeling in \ours is highly challenging to state-of-the-art LMs, where GPT-3.5 almost fails completely, GPT-4 can only generate exactly matching \dfs 16\% of the time and solvable \pfs 33\% of the time, and GPT-4o demonstrate 18\% \dfs accuracy and 37\% \pfs solving rate. By devising a ZPD instruction that prompt LMs to modularly generate PDDL through extraction-inference-translation approach, we improve action accuracy by 3\% and problem solving by 2-7\% . In our analysis, the syntactic errors indicate LMs' weakness in generating low-resource and domain-specific programming languages \cite{cassano2023knowledge} like PDDL, while the semantic errors suggest LMs' inaccuracies to reason about actions and environments.

\begin{table}[]
\centering
\small 
\begin{adjustbox}{max width=\columnwidth}
\begin{tabular}{lll}
\toprule
                    & \#\df                          & Datasets                                                  \\ \midrule
Ours                      &27                              &  \ours                                             \\
\cite{wong2023learning}        &2                                               & MineCraft, ALFRED                               \\
\cite{lyu2023faithful}        &1                                           & SayCan                                             \\
\cite{xie2023translating}        &2                                                   & Blocksworld, ALFRED                    \\
\cite{liu2023llmp}             &7                                  & Blocksworld, etc.                               \\
\cite{huang2023grounded}        &1                            &  Tabletop                 \\
\cite{huang2022language}         &1                           &  VirtualHome                                    \\
\cite{silver2022pddl}           &18                         &  Blocksworld, etc.                                  \\
\cite{valmeekam2022large}        &2                            &  Blocksworld, Logistics                                        \\ \bottomrule
%\cite{lyu-etal-2021-goal}                  & \multicolumn{1}{l}{NL plan} & \multicolumn{1}{l}{WikiHow}      & \multicolumn{1}{l}{Open}             \\
%\cite{sakaguchi-etal-2021-proscript-partially} & \multicolumn{1}{l}{NL plan} & \multicolumn{1}{l}{proScript}    & \multicolumn{1}{l}{Open}             \\ \bottomrule
\end{tabular}
\end{adjustbox}
\caption{Our work proposes and evaluates models using \ours which is open-domain and based on procedural texts, while past work has relied on closed-domain benchmarks which can be expressed with a singular \df with a fixed set of actions, based on some simulation.}
\vspace{-2ex}
\label{tab:related_work}
\end{table}

\section{Task Formulation}
%\niket{Open domain planning: Problem formulation}

The task of predicting a planning domain definition in a text-based environment can be seen as translating natural language texts to PDDL symbolic language, which consists of a domain file (\df) and one or more problem files (\pfs).\\
A \df defines all actions in the environment:
\begin{itemize}[topsep=-2ex,itemsep=-1ex,partopsep=1ex,parsep=1ex,leftmargin=*]
    \item parameters (e.g., water, pot) as a list of typed variables
    \item preconditions (e.g., water and pot belongs to player; water is not treated) as a conjunctive normal form of predicates
    \item effect (e.g., water is treated) as a conjunctive normal form of predicates
\end{itemize}
\iffalse
\begin{itemize}[topsep=-2ex,itemsep=-1ex,partopsep=1ex,parsep=1ex,leftmargin=*]
    \item a header $H$, which consists of
    \begin{itemize}[topsep=-2ex,itemsep=-1ex,partopsep=1ex,parsep=1ex,leftmargin=*]
        \item types of entities (e.g., \textit{object}, \textit{location}, \textit{player})
        \item predicates (e.g., if object is \textit{at} a location)
        \item names of possible actions (e.g., \textit{boil water})
    \end{itemize}

    \item definitions of actions $A$, which consist of
    \begin{itemize}[topsep=-2ex,itemsep=-1ex,partopsep=1ex,parsep=1ex,leftmargin=*]
        \item parameters (e.g., water, pot) as a list of typed variables
        \item preconditions (e.g., water and pot belongs to player; water is not treated) as a conjunctive normal form of predicates
        \item effect (e.g., water is treated) as a conjunctive normal form of predicates
    \end{itemize}
\end{itemize}
\fi
\vspace{+2ex}
A \pf defines the initial and goal environments:
\begin{itemize}[topsep=-2ex,itemsep=-1ex,partopsep=1ex,parsep=1ex,leftmargin=*]
    % \item entities and their type (e.g., rainwater is water)
    \item initial states (e.g., bucket is empty)
    \item goal states (e.g., bucket is filled with rainwater; rainwater is treated)
\end{itemize}
\vspace{+2ex}
We say that a \df and a \pf can be solved if there exists a sequence of actions $A_1,\dots,A_n$ that results in a transition from the initial state to the goal state. 

Traditionally, the task of text-based PDDL generation involves predicting \pf based on text \txt, where a successfully generated \pf can be solved by the predefined \df. 
% action modeling involves predicting \df based on text \txt, where a successfully generated \df can thus solve provided \pfs defined accordingly. 

In this paper, we address an alternative formulation, action modeling ($A$), in which the generated \df, given text \txt and the domain header $H$\footnote{The domain header includes types, predicates, and names of actions in \df. As the information specified by $H$ is guaranteed to be consistent with that of the \pfs, the evaluation is well-defined.}, is capable of producing plans for \pfs.

\section{Dataset}
We introduce the \ours dataset of 27 different \txt-\df-\pfs tuples, drawing procedural texts from wikiHow articles of various topics (see Appendix~\ref{app:topics}). A class of graduate students in a U.S. university with prior knowledge of PDDL are each given a wikiHow article and annotate a \df and multiple corresponding \pfs from the article, each with a gold plan to solve it. On average, there are 13.33 defined actions in a \df and 8.07 instantiated actions in a gold plan. 
% We partition the 27 examples into a 16:5:6 train-development-test splits. In this work, the split is not used and all \dfs are evaluated, as all our methods are without task specific model training. 
% either zero-shot or manually written few-shot approaches without fine-tuning.
% only the development set is used for error analysis; the test set is strictly held out for evaluation. 
In this work, all our data is used for evaluation, as all our methods are without task specific model training. 
Some sample data of \ours can be found in Appendix~\ref{app:sample}. 
% On average, it takes several hours to train each human annotator, and another several hours to produce a \txt-\df-\pfs tuple.
% During prediction, we treat \txt as an annotated one-line summary of all annotated actions, though results using more variations of \txt can be found in Appendix~\ref{sec:ablation_text}. 

\section{Methodology}
%\niket{Split the method into methodology and implementation. Motivate LLMs because NO data exists and expensive to get training data.}
We first introduce a novel prompt design option, ZPD, and then discuss the choices of text format (\txt), which can range from 10 to 2,000 tokens and influence the selection of LMs.
% We provide some first attempts towards our formulation of action modeling in open-domain texts. 
% \vspace{-2ex}
\subsection{ZPD Prompt Design}
% \vspace{-2ex}
% We focus on zero-shot prompting LMs as open-domain PDDL is extremely costly to annotate, leading to hardly any training data. 
To predict domain actions $A$ based on relevant \txt and the header $H$, we prompt an LM in zero-shot or few-shot instructions.
% manner \cite{brown2020language} by describing the task and providing the input. Note that few-shot prompting is prohibitively costly due to the excessive length of a resulting \df. 
Our instruction employs Zone of Proximal Development (ZPD) theory proposed for human learning \cite{vygotsky1978mind}, which is a variant of the chain-of-thought (CoT) approach. In typical CoT, a task is decomposed into several constituents  (steps), i.e., parameters, precondition, and effect. In contrast, according to ZPD, the complex \ours task is decomposed into atomic \textbf{skills}: 1) \textit{extracting} the relevant \textit{description} of an action; 2) \textit{extracting} and \textit{inferring} the incorporated \textit{entities} and their \textit{state changes}; and 3) \textit{translating} the entity-state changes to accessible PDDL predicates. Next, we establish the relationships between these atomic skills: to perform the task, each skill is a prerequisite for the next. Finally, we explicitly instruct the LMs to incrementally perform the three basic skills, leading to the successful completion of the \ours task (the prompt can be found in Appendix~\ref{app:prompts}):

% Our ZPD instruction explicitly guides an LM to gain incremental task abilities through three sub-tasks to solve the more complex \ours task 

% \cite{wei2023chainofthought}
\begin{enumerate}
[topsep=-2ex,itemsep=-1ex,partopsep=1ex,parsep=1ex,leftmargin=*]
\item Extraction: describe each action, including the expected preconditions and effects;
\item Inference: list the involved entities and their state changes;
\item Translation: based on the information above, convert \txt to PDDL.
\end{enumerate}

%See our prompt and example in Appendix \ref{app:prompts}.
%First, \textbf{Identification} of the actions from relevant steps in a wikiHow article.\\
%Second, \textbf{Extraction} of entities and their states from those steps. This also requires inference on implicit entity states (e.g., a cloth gets wet if soaked). Entity states are represented using natural language.\\
%Third, \textbf{Translation} of said entity states to PDDL. This requires paraphrasing of entity states in NL to given predicates in SL (e.g., a cloth getting soaked might translate to \texttt{(submerged ?cloth)}).
%This chain-of-thought formulation provides the model a scaffolding to achieve the task, and makes it more attentive to the entities and changes in entity states, even implicit ones (e.g., a soaking action causes an entity to go from dry to wet). Thus, the model is facilitated to reason about the conditions more completely.

\begin{table}[t!]
\centering
\small
\resizebox{0.8\linewidth}{!}{
% \begin{tabular}{llll}
% \toprule
%             & \multicolumn{1}{c}{Intrinsic} & \multicolumn{2}{c}{Extrinsic} \\ 
%  Model    \%   & action acc.                  & \pf solve  & exact plan \\ \midrule
%  \texttt{gpt-3.5}    & 0.2                         & 1.0       & 1.0         \\ \addlinespace
%  \texttt{gpt-4}       & 15.9                         & 33.7       & 4.2         \\
%   + CoT& 9.3& 21.1&3.2\\
%   + ZPD  & \textbf{18.1}                         & \textbf{35.8}       & \textbf{6.3}         \\
%   + ZPD, 3 shot & 11.9& 23.2& 5.3\\ \addlinespace
%  \texttt{gpt-4o}       & 18.2                         & 37.9       & 5.3         \\
%   + CoT & 19.5& 33.7&6.3\\
%   + ZPD  & \textbf{21.4}                         & \textbf{45.3}       & \textbf{6.3}         \\ 
%  + ZPD, 3 shot& 20.3& 40.0& 4.2\\ \addlinespace
%  gold & 100                             & 100            &  100            \\ \bottomrule
% \end{tabular}
\begin{tabular}{lcc}
\toprule
        & \multicolumn{1}{c}{Intrinsic} & \multicolumn{1}{c}{Extrinsic} \\ 
 Model    \%   & action acc.                  & \pf solve \\ \midrule
 \texttt{gpt-3.5}    & 0.2                         & 1.0      \\ \hline
 \texttt{gpt-4}       & 15.9                         & 33.7    \\
  \texttt{+ CoT}& 9.3& 21.1\\
  \texttt{+ \textbf{ZPD}}  & \textbf{18.1}                         & \textbf{35.8}        \\
  \texttt{+ ZPD, 3 shot} & 11.9& 23.2\\ \hline
 \texttt{gpt-4o}       & 18.2                         & 37.9  \\
  \texttt{+ CoT} & 19.5& 33.7\\
  \texttt{+ \textbf{ZPD}}  & \textbf{21.4}                         & \textbf{45.3}   \\ 
 \texttt{+ ZPD, 3 shot} & 20.3& 40.0\\ \hline
 gold & 100                             & 100      \\ \bottomrule
\end{tabular}
}
\caption{The intrinsic and extrinsic evaluation results for all main models. \texttt{gpt-4(o)} demonstrates non-trivial performance. With a ZPD instruction, the performance improves consistently.}
% \vspace{-2ex}
% \caption{LMs face significant challenges with \ours. The performance of \texttt{gpt-3.5} is nearly zero, while  The experiments presented in the table are conducted on \txt = \texttt{sum}.}
\label{tab:results}
\end{table}
% \niket{ablations can move to appendix and add a sentence about them, T=map, sum, rel. Add chatgpt result.}

\begin{table}[]
\centering
\small
\resizebox{0.8\linewidth}{!}{
\begin{tabular}{lccc}
\toprule
Model \% & Parameter & Precondition & Effect \\ \midrule
\texttt{gpt-4} & 36.7 & 31.1 & \textbf{53.0} \\
 \texttt{+ CoT}& 29.7& 25&54.7\\
\texttt{+ ZPD} & \textbf{42.2} & 29.7 & 48.1 \\ \hline
\texttt{gpt-4o} & 45.1 & 31.1 & \textbf{62.5} \\
\texttt{+ CoT} & 52.4 & 34.2 & 54.1 \\
\texttt{+ ZPD} & \textbf{53.5}& \textbf{40.1}& 53.5\\\bottomrule
\end{tabular}}
\caption{The generation accuracy of each component in actions has been evaluated. The ZPD instruction clearly aids in identifying implicit parameters (entities). Predicting preconditions is more challenging than predicting effects, as it requires a greater depth of implicit knowledge of entity states.}
\vspace{-2ex}
\label{tab:results_intrinsic}
\end{table}

%\niket{Consider adding analysis about marginalizing over subsections of the domain file}

\subsection{Choice of Input Text} \label{sec:ablation_text}
We also consider the following choices of wikiHow text as \txt.\\

\minisection{Prompt without text (w/o \txt)} is an ablation baseline where the model predicts $A$ solely based on $H$. Naturally, none of the three aforementioned stages are involved in this prompt condition.\\

\minisection{Prompt with text (w/ \txt)} additionally provides the model with four different portions of \txt, involving the three aforementioned stages, as follows:\\
% [topsep=-2ex,itemsep=-1ex,leftmargin=*]
\textbf{(\txt= all)}: All steps in a wikiHow article.\\
\textbf{(\txt= rel)}: In \ours, each wikiHow article consists of step paragraphs that may or may not be used in defining the actions in the \df. Hence, a mapping between actions and steps is also annotated. This context includes relevant steps to all actions in a \df. (e.g., Step 1. Find fresh water... Step 2. Collect food... Step 7. Set up camp...)\\
\textbf{(\txt= map)}: Each action is mapped with steps based on the annotated mapping in \ours.\\
(e.g., clean\_water: Step 1. Find fresh water...)\\
\textbf{(\txt= sum)}: An one-line summary of each action annotated in \ours.\\
(e.g., clean\_water; boil water to clean it)\\
The four prompts are increasingly general. Distinguishing from the required skills, the full text condition demands accurate information extraction, while the text summary clearly defines the action but requires the model's robust ability to infer implicit entity states. All prompts request an exact translation.

\begin{table}[t!]
\centering
\small
\resizebox{0.8\linewidth}{!}{
% \begin{tabular}{llll}
% \toprule
%             & \multicolumn{1}{c}{Intrinsic} & \multicolumn{2}{c}{Extrinsic} \\ 
%  Model     \%   & action acc.                  & \pf solve  & exact plan \\ \midrule
%  w/o \txt (baseline)    & 13.7                         & 26.3       & 3.2         \\
%  \txt=sum       & 15.9                         & 33.7       & 4.2         \\
%  \txt=sum, ZPD  & \textbf{18.1}                         & \textbf{35.8}       & \textbf{6.3}         \\
%  \txt=map       & 11.8                         & 13.7       & 2.1         \\
%  \txt=map, ZPD  & 8.9                          & 26.3       & 1.1         \\
%  \txt=rel       & 11.6                         & 27.4       & 0.0         \\
%  \txt=rel, ZPD  & 12.2                         & 21.1       & 4.2         \\
%  \txt=all      & 12.1                         & 28.4       & 0.0         \\
%  \txt=all, ZPD & 12.1                             & 31.6             &  0.0            \\ \bottomrule
% \end{tabular}
\begin{tabular}{lcc}
\toprule
            & \multicolumn{1}{c}{Intrinsic} & \multicolumn{1}{c}{Extrinsic} \\ 
 Model     \%   & action acc.                  & \pf solve \\ \midrule
 w/o \txt (baseline)    & 13.7                         & 26.3  \\
 \txt=sum       & 15.9                         & 33.7  \\
 \txt=sum, \texttt{ZPD}  & \textbf{18.1}                         & \textbf{35.8}   \\
 \txt=map       & 11.8                         & 13.7  \\
 \txt=map, \texttt{ZPD} & 8.9                          & 26.3  \\
 \txt=rel       & 11.6                         & 27.4 \\
 \txt=rel, \texttt{ZPD}  & 12.2                         & 21.1  \\
 \txt=all      & 12.1                         & 28.4  \\
 \txt=all, \texttt{ZPD} & 12.1                             & 31.6 \\ \bottomrule
\end{tabular}
}
\caption{Performance of GPT-4 using different portions of text \txt. Metrics include action-wide accuracy and the proportion of \pfs that can be solved.}
% average edit distance of action definitions, , and the proportion of generated plans that exactly match the gold plans
\vspace{-2ex}
\label{tab:ablation_results}
\end{table}

\subsection{Experiments}
 We conducted experiments with three large language models\footnote{Due to the need for very long input and output, the choice of open-source models is limited. We are in progress of implementing \texttt{Mixtral-8x7B}.}: GPT-3.5-turbo-16k, GPT-4-32k (dated June 2023), and GPT-4o. For GPT-4-32k, we used a maximum token limit of 10,000.
 % and a temperature setting of 0.7. 
 GPT-3.5-turbo-16k and GPT-4o were tested with theirs default hyperparameters. The few-shot examples can be found in Appendix~\ref{app:few_shot}.

\section{Evaluation and Analysis}
Now that a model generates the parameters, preconditions, and effects for each action, we have a complete \df. We evaluate it in two ways (Figure~\ref{fig:formulation}). \textit{Intrinsically}, we semantically compare the predicted $A$ with the ground-truth provided by our \ours and report an action-wide accuracy. Equivalence of two action definitions does not depend on the naming of variables nor on the order within conjunctions (detailed in Appendix \ref{sec:action_equivalence}). \textit{Extrinsically}, to measure actions' coherence, a BFS-based PDDL solver\footnote{\url{https://github.com/pucrs-automated-planning/pddl-parser}} attempts to solve ground-truth \pfs with the predicted \df and a success rate is reported. An unsolved \pf is caused by (1.) no plan can be found, or (2.) the solver runs for more than 30 seconds, or (3.) the solver returns an error (usually a syntax error in the generated PDDL).

%(see Appendix~\ref{sec:action_equivalence})

%\subsection{Evaluation}
The intrinsic and extrinsic results are shown in Table~\ref{tab:results}. \texttt{gpt-3.5-turbo} which achieves impressive performance on many tasks has a close-to-zero performance. In contrast, \texttt{gpt-4} performs significantly better with 18\% action prediction accuracy and 36\% solve rate of \pfs. The most advanced \texttt{gpt-4o} presents the highest performance, with 21\% action accuracy and 45\% \pfs solving rate. Still, the performance is far worse than ideal, showing that even a simplified open-domain planning formulation is challenging to state-of-the-art LMs. \\

\minisection{ZPD Instruction Analysis} \\
ZPD is helpful in each setting since it explicitly spells out many implicit entities and state changes in the inference stage which are critical to predicting parameters. 
In most situations, the model summarizes the action and extracts the entity states correctly, though sometimes missing a few implicit entities. However, ZPD's bottleneck lies in the translation stage, during which there are mainly three types of errors. 
\begin{enumerate}[topsep=-2ex,itemsep=-1ex,partopsep=1ex,parsep=1ex,leftmargin=*]
    \item mismatched predicates: the model uses \texttt{(at ?loc ?item)} instead of \texttt{(inventory ?item)};
    \item hallucinated predicates: the model creates a new predicate \texttt{(soaked ?item)} while neglecting the existing \texttt{(submerged ?item)};
    \item complicated predicates: the model adds unnecessary predicates \texttt{(inventory ?submerged\_item - item)} when already has \texttt{(inventory ?item)}.
\end{enumerate}
\vspace{+1em}
To address these, we leave to future work to demonstrate and standardize the translation process by clearly describing all necessary entity-state change and encouraging the model to compare and strictly match the given predicates. 
%We also notice that the CoT leads to omitted actions facing longer inputs and outputs. To maintain the simplicity of each task, an obvious next step is to separate our joint model into a two-episode pipeline: first summarize action, entity, and states, then translate into PDDL under proposed guidance.
% Howeverun, even with CoT, there are cases of identified entity states being ignored in the translation stage, likely because of the task complexity. and the brevity and coherence of NL text,  %We ask model to summarize most relevant information first (action description, entities and states), Then, giving this succinct output as the translation input, model can focus on generating the correct PDDL output.
Finer-grained evaluation results are shown in Table~\ref{tab:results_intrinsic} to tease out the performance regarding such component within an action. It is clear that the LM is worse at predicting preconditions than at predicting effects. This is understandable as procedural texts like wikiHow tend to be less explicit about predictions than about effects (e.g., from `bake for 10 minutes' it is obvious that the food will be baked, but it is unclear what state it had been in).\\

\minisection{Text Format Analysis}\\
As shown in Table~\ref{tab:ablation_results}, in w/o \txt setting, fully relying on its implicit knowledge, the model is already capable of inferring PDDL syntactically and semantically. In w/ \txt settings, our model shows an `U' performance in terms of the text length. Using a sentence-long description for each action (\txt = \texttt{sum}) provided by \ours, the model achieves the best performance among all, showing a strong deduction ability with the limited but precise NL input. The \txt = \texttt{all} setting ensues, which requires the most extraction rather than inference. In contrast, the middle ones (\txt = \texttt{rel/map}) with decreasing signal-to-noise ratio lead to worse results, indicating its shortage of extraction-inference trade-off. The signals contain both the described entity states and step relations, explicitly and implicitly. This shortage may come less from the entity states \textit{(e.g., fish, spear in hunt\_fish)}, but more from the relation between actions \textit{(e.g., make\_spear to hunt\_fish)} which may be expressed in the \txt = \texttt{sum} and \texttt{all} settings.\\

\begin{table}[t!]
    \centering
    \small
    \begin{tabular}{l|ccc|cc}
    \toprule
           & \multicolumn{3}{c|}{Unsolved}      & \multicolumn{2}{c}{Solved} \\ 
           & \makecell{Syntactic\\Error} & \makecell{Bad\\Action} & \makecell{Good\\Action} & \makecell{Bad\\Plan}    & \makecell{Good\\Plan}    \\ \midrule
\texttt{gpt-4}  & 3      & 7          & 2           & 0           & 3            \\ \bottomrule
%\txt=all & 0      & 10         & 0           & 3           & 2            \\ 
    \end{tabular}
    \caption{A small-sample inspection shows that models make both syntactic and semantic errors.}
    \vspace{-2ex}
    \label{tab:error_analysis}
\end{table}

\minisection{Case Analysis}\\
To provide deeper insights into model performance, we manually inspect the model output of \texttt{gpt-4} on all 6 examples (15 \pfs) in the development set. We consider the following scenarios.

\minisection{Unsolved} Whenever the predicted \df cannot solve a \pf, either a syntactic or a semantic error has occurred. For a \textbf{syntactic error}, the output may contain illegal expressions that cannot be parsed. For example, \texttt{(inventory ?player (clean ?strips))} is unacceptable because the arguments to a predicate must be atomic types, not another predicate. For a \textbf{semantic error} (namely, a `bad action'), we identify the first problematic action that differs with the ground-truth. For example, if the action \texttt{cut\_plant} misses a critical effect of \texttt{(inventory ?player ?stalk)}, then other actions such as \texttt{graft\_stalk} requiring it cannot be executed. At times, there could be false negatives where the predicted action definitions are in fact reasonable but still cannot lead to a solution (namely, a `good action').

\minisection{Solved} Even when the predicted \df solves a \pf, the plan may be different from the gold plan. It is naturally possible that the predicted plan is a fluke made possible by under-specified preconditions or over-exaggerated effects, as well as loopholes in the \pf leading to unreasonable shortcuts. For the example in Figure~\ref{fig:pddl_example}, a model could \textit{cheat} by defining the action \texttt{get} by not requiring the person and object to be in the same location; thus, the predicted plan would unreasonably omit the action \texttt{go}. However, at times, the predicted plan could also be a reasonable alternative.

The statistics of these errors are shown in Table~\ref{tab:error_analysis}. When no solution can be found, true negative is highly likely as the model indeed makes aforementioned mistakes during action prediction. When some solution is found, false positive is still possible as the predicted plan may be unreasonable. See attached materials for a complete error analysis of these examples. Our aforementioned future pipeline that separates summarization and translation would likely mitigate these errors.

\section{Conclusion}
% \vspace{-2ex}
We present \ours, the first open-domain dataset that juxtaposes natural language and planning domain definition language. Our experiments show that ZPD instructions facilitate LMs' performance, while still find it challenging to translate the precondition and effects of actions. We hope our instruction design, evaluations and dataset help future progress towards integrating the best of LM and formal planning.

% \corr{Future directions include 1. developing better models and prompts; 2. improving evaluation to address subjectivity and false negatives; 3. explore more ambitious settings such as predicting full \pf and \df.}

\section{Limitations}

Any planning language, including PDDL which we consider in this work, is an approximation of planning in the real world and cannot accurately reflect its complexity. Due to the consideration for simplicity in the annotation process, we use the primitive version of PDDLs, with restricted expressions and syntax, instead of newer versions of the planning language which extend its syntax in a variety of way. 

Annotating \ours is extremely costly as it requires knowledge of PDDL and much effort to translate procedural texts to PDDL. Thus, our dataset is relatively small with a limited range of topics. Due to the highly complex and subjective nature of the annotation process, each annotated example may reflect idiosyncratic though processes and biases of the individual annotator. 

As with many similar works, there is a known gap between high-level planning such as ours (with high-level actions like ``boil'') and the actions used by present-day robots (with low-level motor functions like ``move''). However, like similar works, we believe our efforts can see more practical application in the near future.

Our modeling efforts so far have mainly considered options of zero-shot prompting. There of course exists many other approaches including the few-shot setting, fine-tuning, and the  model distillaion paradigm, which we plan to experiment with in the future. Moreover, our evaluation is imperfect in that even a well-annotated DF-PF pair might have multiple successful plans. Manual inspection is still necessary to accurately gauge models.

\section*{Acknowledgements}
This research is supported in part by the Office of the Director of National Intelligence (ODNI), Intelligence Advanced Research Projects Activity (IARPA), via the HIATUS Program contract \#2022-22072200005. The views and conclusions contained herein are those of the authors and should not be interpreted as necessarily representing the official policies, either expressed or implied, of ODNI, IARPA, or the U.S. Government. The U.S. Government is authorized to reproduce and distribute reprints for governmental purposes notwithstanding any copyright annotation therein.

\bibliography{custom}

\appendix

\section{Topics}
\label{app:topics}

Below are a list of the titles of wikiHow articles in \ours, chosen per the requirement of a gruaduate-level university class.
\begin{itemize}[noitemsep,topsep=0pt]
    \item create secret society
    \item throw an anime party
    \item open a coconut
    \item calculate pi
    \item hack
    \item get out of quicksand
    \item make a detective kit
    \item lock picking
    \item make papyrus
    \item survive on a desert island
    \item survive in the jungle
    \item survive a war
    \item survive a comet hitting earth
    \item survive a nuclear attack
    \item survive in the woods
    \item survive deserted island
    \item survive shark attack
    \item survive emp attack
\end{itemize}
Each topic may have one or more annotated \dfs representing different domains. The homogeneity of the last 7 topics is due to the class' topic of interactive fictions.

\section{Sample Data: \txt, \df, and \pf}
\label{app:sample}

To exemplify \ours, below is an example procedural text \txt titled `survive in the jungle`, up to the third step, truncating the rest. 

\begin{lstlisting}
1. Collect rainfall from leaves and bamboo stalks. Look for large leaves that collect rainfall and bend them into a funnel to pour the water into a bottle or straight into your mouth. Bend bamboo stalks to let the water that collects in the compartments flow out into a container or break the bamboo compartment off at the line that goes across the stalk to use it as a water bottle. You could also look for rock formations that form natural pools and collect rainwater, but it is best to do this after a fresh rainfall to avoid pools that have been sitting for a long time and may be contaminated with bacteria. If you don't have a water bottle or other container to collect water, try to find other natural containers in the jungle such as a coconut shell or piece of wood shaped like a bowl. You can also leave these items out when it rains to collect the fresh water.

2. Boil water from streams to kill any bacteria. Look for running streams to find fresh water. Filter out any particles through a sock, shirt, or other fabric, then start a fire and boil the water to kill bacteria that can make you sick. If you don't have a pot to boil water in, then you can use a tin can, single-walled stainless steel water bottle, or any other metal container. If you have no way of making a fire or boiling the water, then you should avoid drinking water from streams. It can be contaminated with many types of bacteria from animals that will make you very sick. Always avoid drinking water from stagnant pools as the water is likely contaminated.

3. Make a solar water still with a container and a plastic sheet. Dig a hole in an area that receives at least some direct sunlight and put a container, such as a water bottle or can, in the middle of the hole. Fill the space between the sides of the hole and the container with wet leaves. Place a plastic sheet over the top of the hole and put rocks or other heavy objects around the edges to hold it in place. Put a small stone in the middle of the sheet above the container. The plastic sheet will accumulate condensation that will drip down the underside of the sheet and into the container. This water is distilled and safe to drink. You can use natural containers such as bamboo or a coconut shell if you don't have a bottle or can. A solar still does not collect large amounts of water. It should be used as a supplemental source of water rather than a primary source.

......
\end{lstlisting}

Below is a sample annotated \df of the above:
\begin{lstlisting}
(define (domain survive_in_the_jungle)
   (:requirements :strips :typing)
   (:types
       stone wood bamboo_container water fire sos_sign fruit - item 
       basecamp - location
       ill dehydrated hungry - condition
       player
       direction 
   )

   (:predicates
      (has_bamboo ?loc - location) ; this location has bamboo to create a container
      (has_rainfall ?loc - location) ; this location has received rainfall to collect water
      (has_fruit ?loc - location) ; this location has fruits to pick
      (treated ?water - water) ; True if the water has been decontaimated by boiling it
      (is ?c - condition ?p - player) ; True if the player is under the specified condition
      (at ?obj - object ?loc - location) ; an object is at a location 
      (inventory ?player ?item) ; an item is in the player's inventory
      (connected ?loc1 - location ?dir - direction ?loc2 - location) ; location 1 is connected to location 2 in the direction
      (blocked ?loc1 - location ?dir - direction ?loc2 - location) ; the connection between location 1 and 2 in currently blocked
   )

   (:action go ; navigate to an adjacent location 
      :parameters (?dir - direction ?p - player ?l1 - location ?l2 - location) 
      :precondition (and (at ?p ?l1) (connected ?l1 ?dir ?l2) (not (blocked ?l1 ?dir ?l2)))
      :effect (and (at ?p ?l2) (not (at ?p ?l1)))
   )

   (:action get ; pick up an item and put it in the inventory
      :parameters (?item - item ?p - player ?l1 - location) 
      :precondition (and (at ?p ?l1) (at ?item ?l1))
      :effect (and (inventory ?p ?item) (not (at ?item ?l1)))
   )

   (:action get_bamboo_container; get a bamboo container using surrounding bamboo
      :parameters (?p - player ?loc - location)
      :precondition (and (at ?p ?loc) (has_bamboo ?loc))
      :effect (inventory ?p bamboo_container)
   )

   (:action collect_rain_water
      :parameters (?p - player ?loc - location)
      :precondition (and (at ?p ?loc) (inventory ?p bamboo_container) (has_rainfall ?loc))
      :effect (and (inventory ?p water) (not (treated water)))
   ) 

   (:action create_fire
      :parameters (?p - player ?loc - location)
      :precondition (and (at ?p ?loc) (inventory ?p stone) (inventory ?p wood))
      :effect (and (at fire ?loc) (not (inventory ?p stone)) (not (inventory ?p wood)))
   )

   (:action treat_water
      :parameters (?p - player ?loc - location)
      :precondition (and (inventory ?p water) (not (treated water)) (at fire ?loc))
      :effect (and (treated water))
   )

   (:action drink_water
      :parameters (?p - player)
      :precondition (and (inventory ?p water) (treated water))
      :effect (not (is dehydrated ?p))
   )

   (:action drink_untreated_water
      :parameters (?p - player)
      :precondition (and (inventory ?p water) (not (treated water)))
      :effect (is ill ?p)
   )

   (:action create_sos_sign
      :parameters (?p - player)
      :precondition (and (inventory ?p stone) (at ?p basecamp))
      :effect (and (not (inventory ?p stone)) (at sos_sign basecamp))
   )

   (:action pick_fruit
      :parameters (?p - player ?loc - location)
      :precondition (and (at ?p ?loc) (has_fruit ?loc))
      :effect (inventory ?p fruit)
   )

   (:action eat_fruit
      :parameters (?p - player)
      :precondition (and (is hungry ?p) (inventory ?p fruit))
      :effect (and (not (inventory ?p fruit)) (not (is hungry ?p)))
   )

   (:action escape
      :parameters (?p - player)
      :precondition (and (at ?p basecamp) (at sos_sign basecamp) (not (is hungry ?p)) (not (is dehydrated ?p)) (not (is ill ?p)))
      :effect (not (at ?p basecamp))
   )
)
\end{lstlisting}

Below is an annotated \pf of the above:
\begin{lstlisting}
(define (problem escape)
   (:domain survive_in_the_jungle)

   (:objects
      npc - player
      jungle bamboo_forrest basecamp - location
      in out north south east west up down - direction
      stone wood sos_sign - item
      ill dehydrated hungry - condition
   )

   (:init
      (at npc basecamp)
      (connected basecamp west bamboo_forrest)
      (connected bamboo_forrest east basecamp)
      (connected basecamp east jungle)
      (connected jungle west basecamp)

      (has_bamboo bamboo_forrest)
      (has_fruit jungle)
      (has_rainfall jungle)

      (at stone bamboo_forrest)
      (at stone basecamp)
      (at wood jungle)
      
      (is dehydrated npc)
      (is hungry npc)
   )
  
   (:goal (and (not (at npc basecamp)) (not (at npc bamboo_forrest)) (not (at npc jungle)) ))
)
\end{lstlisting}
\section{Few-shot Examples}
\label{app:few_shot}
\begin{lstlisting}
Example 1: 
action: slide_straw_over_skewer

text: Slide a segment of plastic straw over the 2 skewers. Cut a segment of a plastic straw that is equal to the width between the wheel wells on your car. Then, slide it onto 1 of the skewers that's attached to a wheel. Do the same with the other skewer.

 pddl_action: 
 (:action slide_straw_over_skewer
 :parameters (?straw1 - straw ?straw2 - straw ?skewer1 - skewer ?skewer2 - skewer ?wheel1 - wheel ?wheel2 - wheel)
 :precondition (and 
 (attached ?skewer1 ?wheel1)
 (attached ?skewer2 ?wheel2)
 (equal_width ?straw1 ?skewer1)
 (equal_width ?straw2 ?skewer2))
 :effect (and 
 (over ?straw1 ?skewer1)
 (over ?straw2 ?skewer2)))

entity_state_inference: 
- action name: slide_straw_over_skewer
- summarize action: The action involves cutting a segment of a plastic straw and sliding it over two skewers attached to wheels. The straw segments should be equal to the width between the wheel wells on a car.
- what is needed to do the action: A segment of a plastic straw, two skewers attached to wheels.

- what is the result: The straw segments are slid over the skewers, which helps in maintaining the structure of the car and providing stability to the wheels.

- entities:
 - entity name: plastic straw segment
 - before: Not cut and not on skewers
 - after: Cut and slid onto skewers
 - entity name: skewer
 - before: Without straw segments
 - after: With straw segments slid onto them
 - entity name: wheel
 - before: Attached to skewers, without straw segments on the skewers
 - after: Attached to skewers, with straw segments on the skewers

Example 2:
action: mix_sand_with_color

text: Close the bags tightly, then shake and squeeze them to mix the sand. Zip the bags shut tightly first. Next, shake the bags to help distribute the food coloring. You can also gently squeeze and massage the sand through the bags. Keep doing this until the color is even. Add more food coloring if the color is too light for you. Keep in mind that the sand will dry lighter.

pddl_action: 
(:action mix_sand_with_color 
:parameters (?bag - bag ?sand - sand ?color - food_coloring) 
:precondition (and (sealed ?bag) (contains ?bag ?sand) (contains ?bag ?color) (not (evenly_colored ?sand))) 
:effect (evenly_colored ?sand))''',

entity_state_inference: 
- action name: mix_sand_with_color
- summarize action: This action involves mixing colored sand by sealing bags containing sand and food coloring, then shaking, squeezing, and massaging the bags until the sand color is evenly distributed. More food coloring may be added if the color is too light, with the understanding that the sand will dry lighter.
- what is needed to do the action: Sealed bags containing sand and food coloring, ability to shake and squeeze the bags.
- what is the result: The sand inside the bags is evenly colored.

- entities:
 - entity name: bags
 - before: contains uncolored or unevenly colored sand and food coloring
 - after: contains evenly colored sand
 - entity name: sand
 - before: uncolored or unevenly colored
 - after: evenly colored
 - entity name: food_coloring
 - before: present in the bags
 - after: mixed evenly with the sand

Example 3:
action: melt_wax

text: Melt candle wax flakes or cubes in a double boiler. Set a can, jar, or another heat-proof container in a pot, and place your solid candle wax (soy or paraffin are most common) inside. Surround the container with water, filling the pot about halfway up the container, then heat the pot on medium heat to double boil the wax to melt it completely, stirring every minute or so to make sure it melts evenly.

pddl_action: 
(:action melt_wax
 :parameters (?wax ?container ?pot ?heat_source ?water)
 :precondition (and (solid ?wax) (in ?wax ?container) (heatproof ?container) (in ?container ?pot) (in ?water ?pot) (cold ?water))
 :effect (and (liquid ?wax) (heated ?water)))

entity_state_inference: 
- action name: melt_wax
- summarize action: This action involves melting solid candle wax using a double boiler method. The solid wax is placed in a heat-proof container, which is then placed in a pot filled with water. The pot is heated, and the wax is stirred until it melts completely.
- what is needed to do the action: The action requires solid wax, a heat-proof container, a pot, water, and a heat source.
- what is the result: The solid wax is melted into liquid wax.

- entities:
 - entity name: wax
 - before: solid
 - after: liquid
 - entity name: container
 - before: empty or containing solid wax
 - after: containing liquid wax
 - entity name: pot
 - before: empty or containing water and container with solid wax
 - after: containing water and container with liquid wax
 - entity name: water
 - before: cold or room temperature
 - after: heated
 - entity name: heat_source
 - before: off
 - after: on
\end{lstlisting}

\section{Prompts}
\label{app:prompts}

For reproducibility, we provide the verbatim prompts that we used in the above experiments.

% \subsection{Prompt with CoT}\\
\subsection{Prompt without ZPD}
\noindent Could you fill out the below PDDL actions with the predicates based on the text?\\
\noindent All fields: parameters, precondition and effect, should have predicates.\\
\noindent For each action, do NOT change the name and do NOT drop the action and do NOT add more actions.\\
\noindent The output should be in correct PDDL format.\\

\noindent <wikiHow text and domain header>\\\\
\noindent here are the actions I want:\\
\noindent <insert\_action\_names>\\
here are the types I have:\\
\noindent <insert\_types>\\
\noindent here are the predicates I have:\\
\noindent <insert\_predicates>\\
\noindent here are the texts containing steps to
 <insert\_goal>:\\
\noindent <insert\_text>\\

\noindent \textbf{Example Completion:}\\
(:action clean\_water

 :parameters (?player - human ?water - water)
 
 :precondition (inventory ?player ?water)
 
 :effect (treated ?water)\\
)\\

\subsection{Prompt with ZPD}
Could you fill out the below PDDL actions with the predicates based on the text?
All fields: parameters, precondition and effect, should have predicates.\\
For each action, do NOT change the name and do NOT drop the action and do NOT add more actions and:\\
\noindent First, summarize the action in a few sentences based on the text and provide its requirements and its aims if it has.\\
Next, identify ALL the entities involved in the action and describe whether it changed, unchanged, added, removed in the action in natural language.\\
Last, translate it into PDDL format. Check all the related entities are in the 'parameters'.\\

\noindent Please use this output format:\\
- action name: ...\\
- summarize action: ...\\
- what is needed to do the action: ...\\
- what is the result: ...\\

\noindent - entities:\\
  - entity name: ...\\
  - before: ...\\
  - after: ...\\

\noindent - describe how to match it to relevant predicates step by step:\\
1. ...\\
2. ...\\

\noindent <wikiHow text and domain header>\\\\
\noindent here are the actions I want:\\
<insert\_action\_names>\\

\noindent here are the types I have:\\
<insert\_types>\\

\noindent here are the predicates I have:\\
<insert\_predicates>\\

\noindent here are the texts containing steps to <insert\_goal>:\\
<insert\_text>\\

\noindent \textbf{Example Completion:}\\
- action name: clean\_water\\
- summarize action: The player cleans water in their inventory using heat from a fire.\\
- what is needed to do the action: The player must have untreated water in their inventory and be at a location with fire.\\
- what is the result: The player has treated water in their inventory.\\

\noindent - entities:\\
 - entity name: player\\
    - before: Having untreated water in inventory.\\
    - after: Having treated water in inventory.\\
  - entity name: water\\
    - before: Untreated.\\
    - after: Treated.\\

\noindent - describe how to match it to pddl relevant predicates step by step:\\
1. Check if the player has untreated water in their inventory.\\
2. Check if the player is at a location with a fire.\\
3. Replace untreated water with treated water in the player's inventory in the effect.\\

\noindent PDDL:\\
(:action clean\_water

  :parameters (?player - human ?loc - location ?water - water)
  
  :precondition (and (at ?player ?loc) (inventory ?player ?water) (not (treated ?water)) (has\_fire ?loc))
  
  :effect (treated ?water)\\
)\\

\section{Calculating Actions Equivalence}
\label{sec:action_equivalence}
\noindent The distance between two actions can be divided to two parts:
\begin{enumerate}[leftmargin=1em]
    \item The distance between parameters:
    
    We do not need to consider the specific parameter names; we only need to consider the parameter types. For each parameter in Action1, we iterate over the parameter list of Action2 to find the first parameter in Action2 with the same type. We use two hash maps, p1 and p2, to record these two parameters and their corresponding types. We increment the counter by 1, remove that parameter from the parameter list of Action2, and break from the current loop. After the iteration, we obtain the number of matching parameters, n. The distance between parameters can be calculated as $| \text{{number of parameters in Action1}} - n | + | \text{{number of parameters in Action2}} - n |$.
    
    \item The distance between preconditions/effects:
    
    For each condition in Action1, we iterate over the condition list of Action2. The conditions can only match if they have the same predicate and the same number of parameters. We iterate over the parameters in these conditions and make the following judgments:
    
    \begin{itemize}[leftmargin=1em]
        \item If neither of the two current parameters has appeared before (in p1 and p2) and these parameters are not identical, they don't match.
        \item If the two parameters have different categories, they don't match.
        \item If the two parameters have the same categories and don't have an index, we consider them as the same parameter entity and give them the same index. We continue the iteration.
        \item If the two parameters already have indexes, we check if the indexes are equal. If they are not equal, they don't match. Otherwise, we continue the iteration.
        \item In any other case, they don't match.
    \end{itemize}
    
    If all parameters of the two conditions match, we increment n by 1. The distance between preconditions/effects can be calculated as $| \text{{number of preconditions/effects in Action1}} - n | + | \text{{number of preconditions/effects in Action2}} - n |$.
    
\end{enumerate}

\end{document}